\documentclass[sigconf]{acmart}

\AtBeginDocument{%
  \providecommand\BibTeX{{%
    \normalfont B\kern-0.5em{\scshape i\kern-0.25em b}\kern-0.8em\TeX}}}

\setcopyright{acmcopyright}
\copyrightyear{2018}
\acmYear{2018}
\acmDOI{10.1145/1122445.1122456}

\acmConference[Woodstock '18]{Woodstock '18: ACM Symposium on Neural
  Gaze Detection}{June 03--05, 2018}{Woodstock, NY}
\acmBooktitle{Woodstock '18: ACM Symposium on Neural Gaze Detection,
  June 03--05, 2018, Woodstock, NY}
\acmPrice{15.00}
\acmISBN{978-1-4503-XXXX-X/18/06}

\usepackage{graphicx}
\usepackage{amsmath}
\usepackage{amsthm}
\usepackage{booktabs}
\usepackage{algorithm}
\usepackage{algorithmic}
\usepackage{multirow}
\usepackage{epstopdf}
\usepackage{subfigure}

\usepackage{amsmath,amssymb,amsfonts}
\usepackage{algorithmic}
\usepackage{graphicx}
\usepackage{textcomp}
\usepackage{xcolor}

\copyrightyear{2020} 
\acmYear{2020} 
\setcopyright{acmcopyright}\acmConference[CIKM '20]{Proceedings of the 29th ACM International Conference on Information and Knowledge Management}{October 19--23, 2020}{Virtual Event, Ireland}
\acmBooktitle{Proceedings of the 29th ACM International Conference on Information and Knowledge Management (CIKM '20), October 19--23, 2020, Virtual Event, Ireland}
\acmPrice{15.00}
\acmDOI{10.1145/3340531.3411937}
\acmISBN{978-1-4503-6859-9/20/10}

\begin{document}

\title{Controllable Multi-Character Psychology-Oriented Story Generation}

\author{Feifei Xu}
\affiliation{\institution{Shanghai University of Electric Power}}
\email{xufeifei@shiep.edu.cn}

\author{Xinpeng Wang}
\affiliation{%
  \institution{Shanghai University of Electric Power}
}
\email{wangxinpeng@mail.shiep.edu.cn}

\author{Yunpu Ma}
\authornote{The corresponding author}
\affiliation{%
  \institution{Ludwig Maximilian University}}
\email{cognitive.yunpu@gmail.com}

\author{Volker Tresp}
\affiliation{%
  \institution{Siemens Corporate Technology}
  \institution{Ludwig Maximilian University}}
\email{volker.tresp@siemens.com}

\author{Yuyi Wang}
\affiliation{\institution{ETH Zurich, Switzerland}}
\email{yuwang@ethz.ch}

\author{Shanlin Zhou}
\affiliation{%
  \institution{Shanghai University of Electric Power}
}
\email{zhoushanlin@mail.shiep.edu.cn}

\author{Haizhou Du}
\affiliation{\institution{Shanghai University of Electric Power}}
\email{duhaizhou@shiep.edu.cn}

\begin{abstract}
Story generation, which aims to generate a long and coherent story automatically based on the title or an input sentence, is an important research area in the field of natural language generation. There is relatively little work on story generation with appointed emotions. Most existing works focus on using only one specific emotion to control the generation of a whole story and ignore the emotional changes in the characters in the course of the story. In our work, we aim to design an emotional line for each character that considers multiple emotions common in psychological theories, with the goal of generating stories with richer emotional changes in the characters. To the best of our knowledge, this work is first to focuses on characters' emotional lines in story generation. We present a novel model-based attention mechanism that we call SoCP (Storytelling of multi-Character Psychology). We show that the proposed model can generate stories considering the changes in the psychological state of different characters. To take into account the particularity of the model, in addition to commonly used evaluation indicators(BLEU, ROUGE, etc.), we introduce the accuracy rate of psychological state control as a novel evaluation metric. The new indicator reflects the effect of the model on the psychological state control of story characters. Experiments show that with SoCP, the generated stories follow the psychological state for each character according to both automatic and human evaluations. 
\end{abstract}



\maketitle

\fancyhead{}

\section{Introduction}

\begin{table}[htbp]
\caption{An illustration of our story-generation system. The input sentence is the first sentence of the story, and our model appoints the characters like ``Jervis'' and ``Girlfriend'' and three psychological states of this story, like ``Plutchik'', ``Maslow'' and ``Reiss''. For Plutchik, every character has its own emotional lines and will change with our changing setup. The characters will share identical motivations based on Maslow and Reiss. Outputs of the model are the following four sentences that continue the story.
The last box shows an example of the story.}
\centering
\begin{tabular}{|l|l|l|}
\hline
Input Sentence & \multicolumn{2}{l|}{Jervis has been single for a long time.} \\ \hline
Characters & \multicolumn{2}{l|}{Jervis, Girlfriend} \\ \hline
\multirow{2}{*}{Plutchik} & Jervis & sadness--anticipation--joy--joy--joy\\ \cline{2-3}
& Girlfriend & none--none--joy--joy--joy  \\ \hline
Maslow & \multicolumn{2}{l|}{love} \\ \hline
Reiss & \multicolumn{2}{l|}{contact, romance} \\ \hline
\multirow{5}{*}{Story} 
& \multicolumn{2}{l|}{\textbf{Jervis has been single for a long time.}} \\ \cline{2-3}
& \multicolumn{2}{l|}{He wants to have a girlfriend.} \\ \cline{2-3}
& \multicolumn{2}{l|}{One day he meets a nice girl at the grocery store.} \\\cline{2-3}
& \multicolumn{2}{l|}{They begin to date.} \\ \cline{2-3}
& \multicolumn{2}{l|}{Jervis is happy that he is no longer single.} \\ \hline
\end{tabular}
\label{tab1}
\end{table}

"Storytelling is our speciality," the historian Yuval Noah Harari claimed in an interview, "It’s the basis for everything we do as a species." And in his famous book, Sapiens: A brief history of humankind, how do human beings can control the planet? -- Through storytelling and gossip, by believing in shared fictions or myths. Even though one could disagree with his view, we can assume that a good storyteller can share her imagination and can use her emotions to infect others. The goal of this paper is to make machines learn to generate stories, a research topic in natural language generation.
Story generation, in contrast to general text generation, requires coherence of texts in agreement with a sequence of events. Story generation is an important but challenging task in natural language
generation, and it has also become one of the testing methods for advances in the field of AI. In addition to requiring coherence, there exist many challenges in story generation, such as thematic consistency, content consistency, word diversity, and sentiment control.
There is considerable prior research on story generation, such as \cite{fan-etal-2018-hierarchical}, \cite{fan-etal-2019-strategies} and \cite{yao2019plan}. 
However, these works focus only on how to generate a coherent, reasonable and diversified story. Recently, several works have developed controllable emotional text generation that can generate a story based on people's expected emotions, such as
\cite{huang-etal-2018-automatic}, \cite{song-etal-2019-generating} and \cite{luo-etal-2019-learning}.
Although these approaches can generate  texts with different sentiments by appointing the sentiment to the story, they can consider only the sentiment of the whole story, and they are unable to control each character’s sentiment in the course of the story.

Different from the above works, we propose a new model called SoCP (\textbf{S}torytelling \textbf{o}f Multi-\textbf{c}haracter \textbf{P}sychology), which generates a story by appointing the psychological states for each character, and assigns multiple emotions and motivations in accordance with specified psychological theories. An example of a generated story is shown in Table \ref{tab1}. It is a difficult task to consider and control multiple characters' psychological state lines simultaneously. 
In addition, psychological state lines normally involve multiple emotions based on the specified psychological theories.
Compared to controlling a single character and the character's emotions in story generation, this is an extremely challenging task. 
To address these problems, we design two important components: the character selector and the psychological state controller. The character selector will select a current character and the corresponding psychological states from the psychological state line that was designed manually for multiple characters in the story and then proceed to the next component. The psychological state controller will accept the psychological states selected from the character selector, and then use an attention mechanism to control which psychological states and how many will be injected into the generating model.
To account for generating a story that considers psychological changes in its characters, our designed model can generate a story by capturing each character's psychological state line even if it is sometimes implied and abstract.

\begin{figure}[htbp]
\subfigure[Maslow (left) and Reiss (right)]{
\begin{minipage}[t]{0.9\linewidth}
\centering
\includegraphics[scale=0.4]{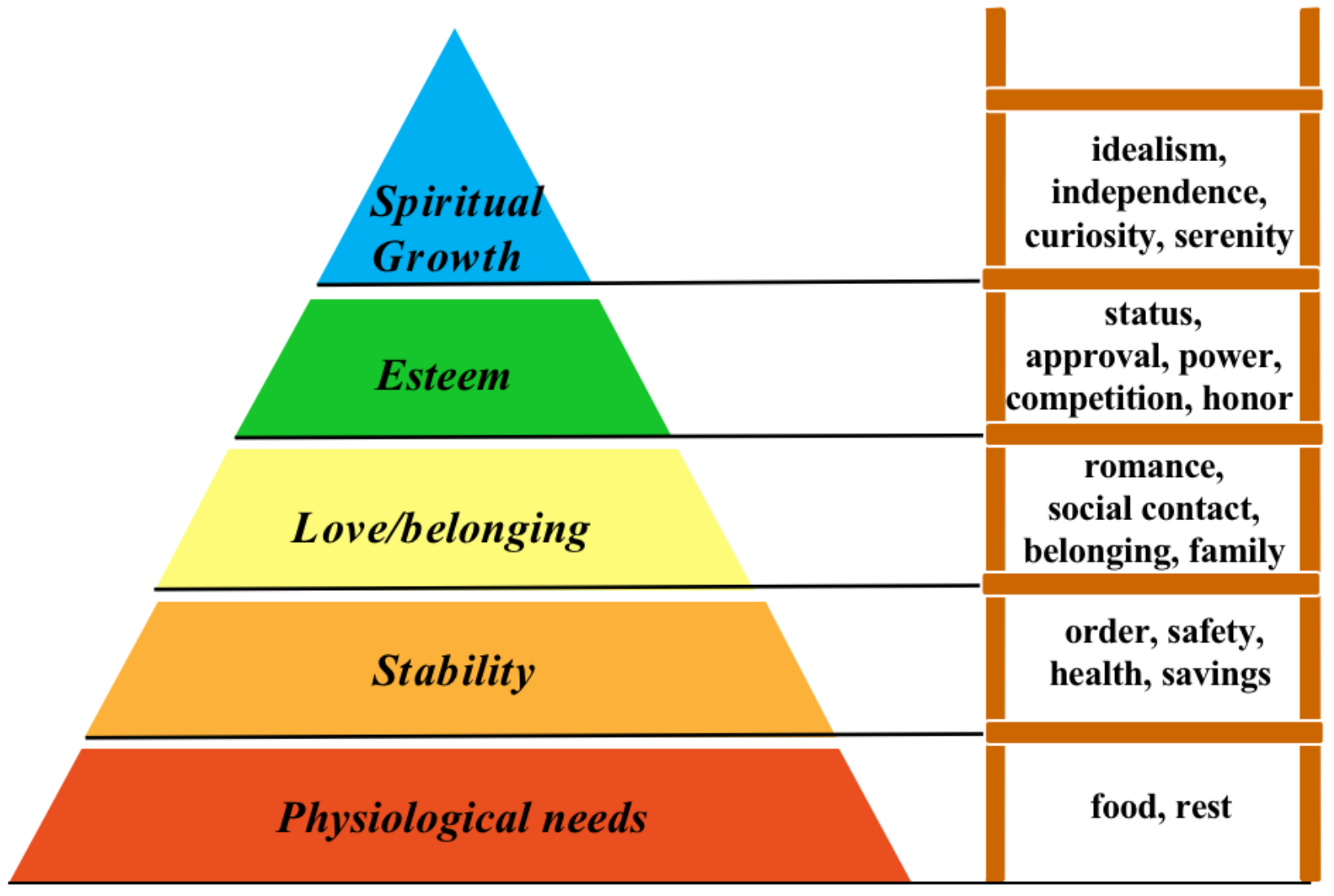}
\label{mr}
\end{minipage}
}

\subfigure[Pluchik]{
\begin{minipage}[t]{0.9\linewidth}
\centering
\includegraphics[scale=0.11]{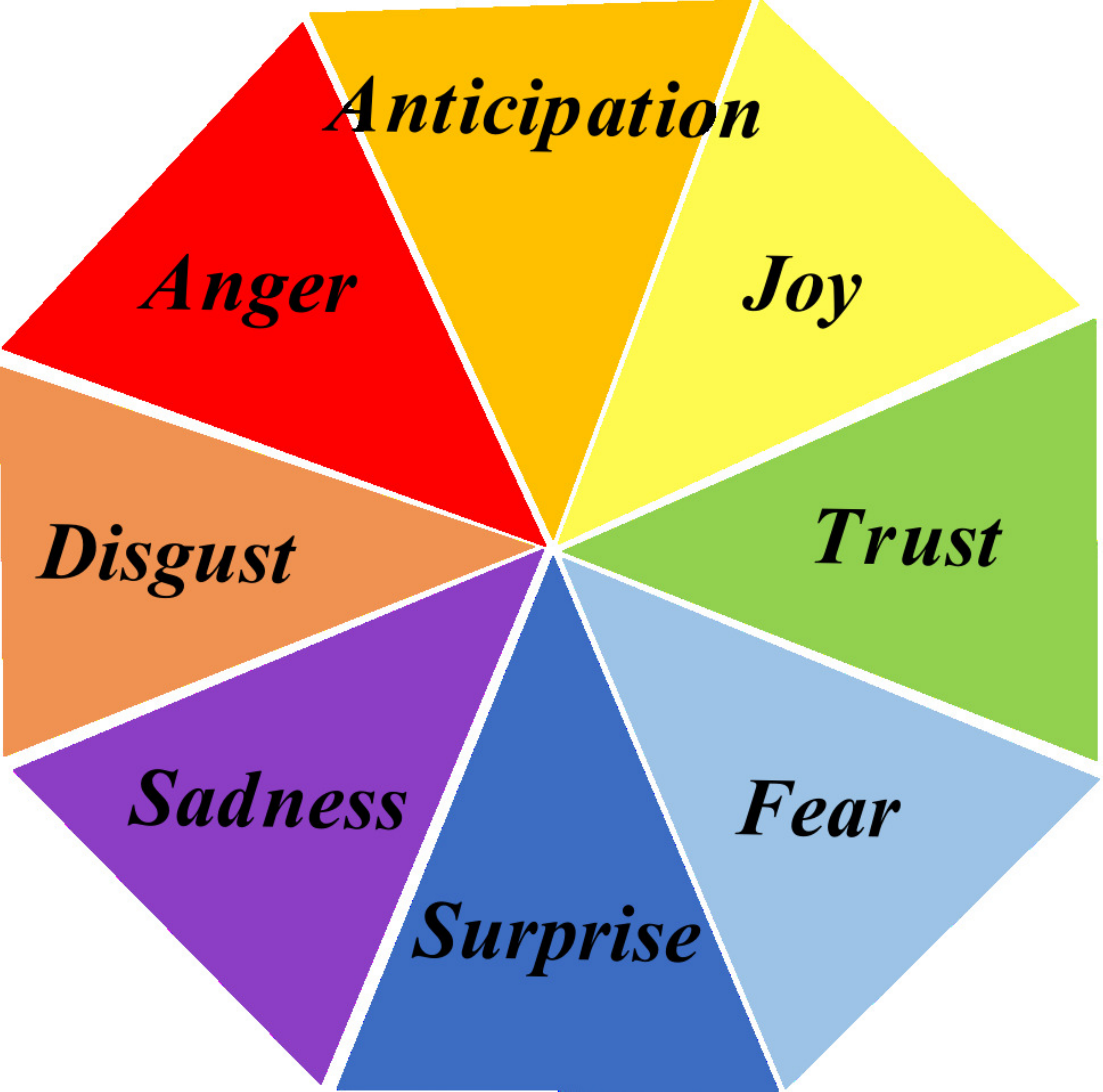} 
\label{p}
\end{minipage}
}
\caption{The psychological states of Maslow, Reiss and Pluchik. The pyramid in (a) shows the classification of Maslow from the physiological to the spiritual level. The right ladder represents the Reiss motives as a subset of Maslow's hierarchy of needs. Plutchik's wheel of emotions in (b) represents the eight basic dimensions of emotions.}
\label{fig1}
\end{figure}

The considered psychological theories are shown in Figure \ref{fig1}.
We use two popular theories to describe the motivation of a person in Figure \ref{mr}: the “hierarchy of needs” of Maslow \cite{maslow1943theory} (left) and the “basic motives” of Reiss \cite{reiss2004multifaceted} (right).
Maslow’s “hierarchy of needs” has five categories from physiological needs to spiritual growth, providing a coarse-level representation. Reiss has 19 fine-grained categories that provide a more informative range of motivations, which can be considered to be a subset of the Maslow's motivations.
The theories of emotion use the “wheel of emotions” of Plutchik \cite{plutchik1980general} in Figure \ref{p}. It has eight basic dimensions of emotions to adequately portray a person.

\begin{figure*}[htbp] 
\centering 
\includegraphics[scale=0.2]{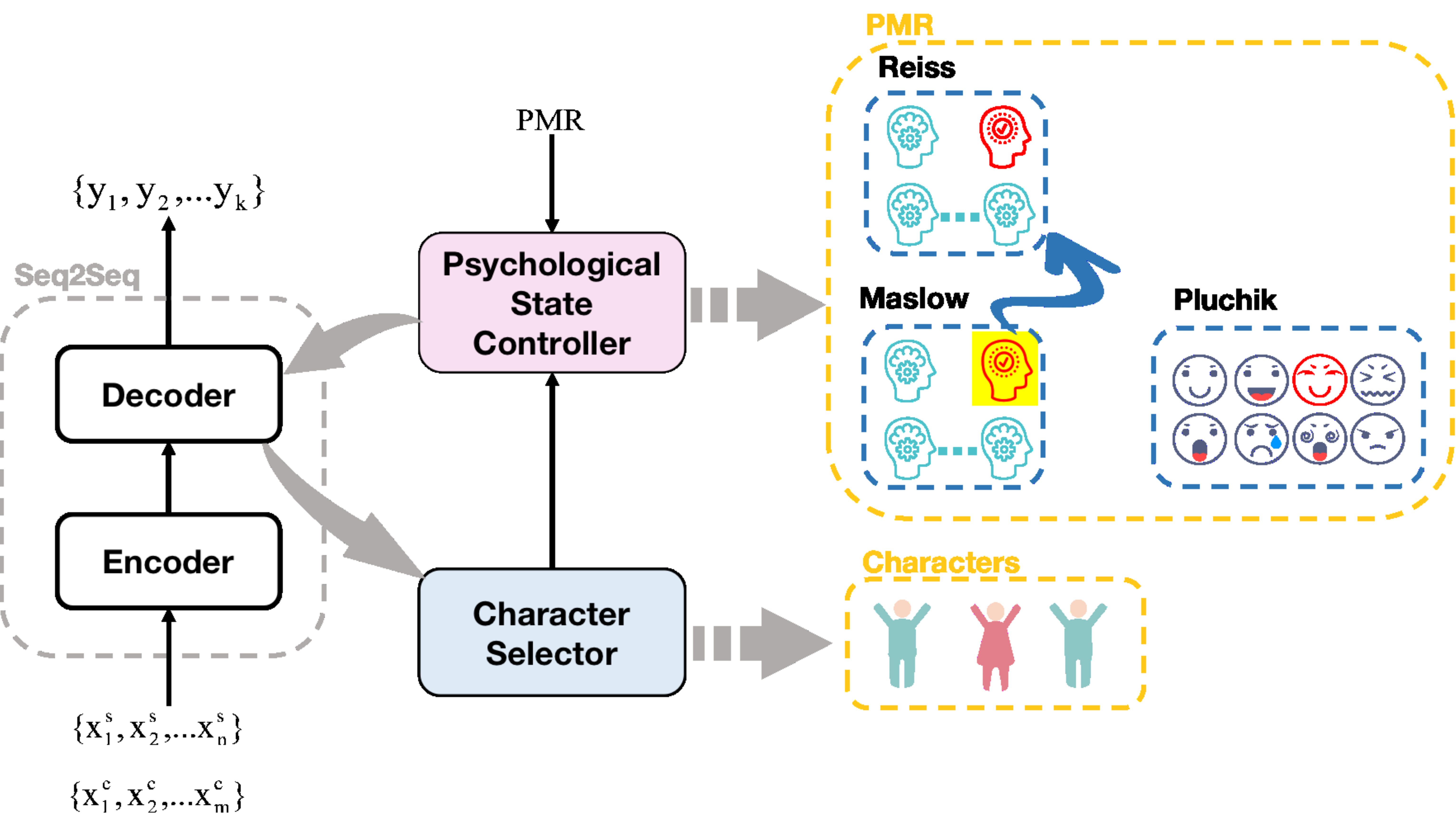} 
\caption{An Overview of the SoCP model, which based on the Seq2Seq structure. The PMR, or psychological state, is the input of the model. Two modules, the Character Selector and the Psychological State Controller, are designed to facilitate the generation of storylines with expected psychological states for each character in the story. The patterns in the yellow dotted line are illustrative descriptions of the two key modules in the model.}
\label{fig2Overview}
\end{figure*}

Our contributions are as follows: 
\begin{itemize}

\item 
We automatically construct the Plutchik matrix, Maslow matrix, and Reiss matrix, which use a matrix mapping method to represent a person's psychological state.

\item 
We propose a model called SoCP that will generate a story with multiple changes in emotion in characters based on the given characters and the corresponding psychological state lines.
To ensure the coherence of the generated story and to achieve multi-character and multi-emotion control, we chose the Seq2Seq framework to generate the story. 
In addition, we design a character selector that determines which character should be selected to be described, and a psychological state controller based on an attention mechanism \cite{bahdanau2014neural} that can control which and how many psychological states the model accepts.

\item 
For evaluation, we develop an accuracy rate of psychological state control as a novel evaluation metric by using a psychology classifier. The experimental results show good performance in various automatic evaluating indicators (BLEU, ROUGE, etc.).
The examples show that our model can generate fluent and reasonable stories based on the dataset.
\end{itemize}

\section{Related Work}

\subsection{Story Generation}
Story generation has attracted much attention over the past years. Recently, many novel ideas have been proposed. For instance, \cite{Jain2017StoryGF} generate a coherent story from independent descriptions, describing a scene or an event. \cite{fan-etal-2019-strategies} explore a strategy for story generation. 
Their frameworks both use sequence-to-sequence neural networks. Previous works have been experimentally proven that sequence-to-sequence neural networks \cite{10.5555/2969033.2969173} have good performance in text generation, e.g., in machine translation \cite{bahdanau2014neural}, summarization \cite{rush-etal-2015-neural}, and dialogue generation \cite{song-etal-2019-generating}.
Considering the outstanding performance of the Seq2Seq structure, our model is designed-based on the sequence-to-sequence model.
Our model is constructed with two specific components: the character selector and the psychological state controller. In addition, multiple emotional inputs based on psychological theory, such as the Plutchik matrix, the Maslow matrix, and the Reiss matrix are considered.

There are also many works using other structures. \cite{fan-etal-2018-hierarchical} use a hierarchical model that first generates a prompt, and then condition on the prompt when generating a story.
\cite{yudraft} propose a multi-pass hierarchical conditional variational autoencoder model to enhance wording diversity and content consistency of the generated story.

\subsection{Text Generation with Emotion}
To generate a story considering our specific psychological theories, we introduce psychological states in addition to the actual sentence. There are also many works in text generation with emotion.
\cite{Xing2016TopicAN} incorporate topic information into a sequence-to-sequence framework to generate informative and interesting responses for chatbots. 
\cite{ghosh-etal-2017-affect} introduce a novel language model for generating affective conversational text conditioned on context words, an affective category and an affective strength parameter. The model can generate expressive text at varying degrees of emotional strength without affecting grammatical correctness. 
Several works have been proposed in the area of emotional or sentimental dialogue generation \cite{huang-etal-2018-automatic,zhou2018emotional,zhou-wang-2018-mojitalk}. They express emotion or sentiment by defining a binary sentiment label (positive/negative) or a set of emotions, such as “anger” and “love”.

Different from these papers, we do not just add emotion to the story but assign different emotion lines to different characters. In particular, these emotions refer to the specified psychological theories, which include three indicators, and each indicator has many emotions or motivations that can describe a character more precisely. Thus, according to the particularity of this task, we choose a dataset from \cite{rashkin-etal-2018-modeling}, which has many fine-grained psychologies for every character in the story. \cite{paul-frank-2019-ranking} used our dataset to extract, rank, filter, and classify sentiments in terms of their underlying human needs.

\section{Model}

\subsection{Overview}
\begin{figure*}[htbp]
\centering
\subfigure[Encoder]{
\begin{minipage}[t]{0.45\linewidth}
\centering
\includegraphics[scale=0.23]{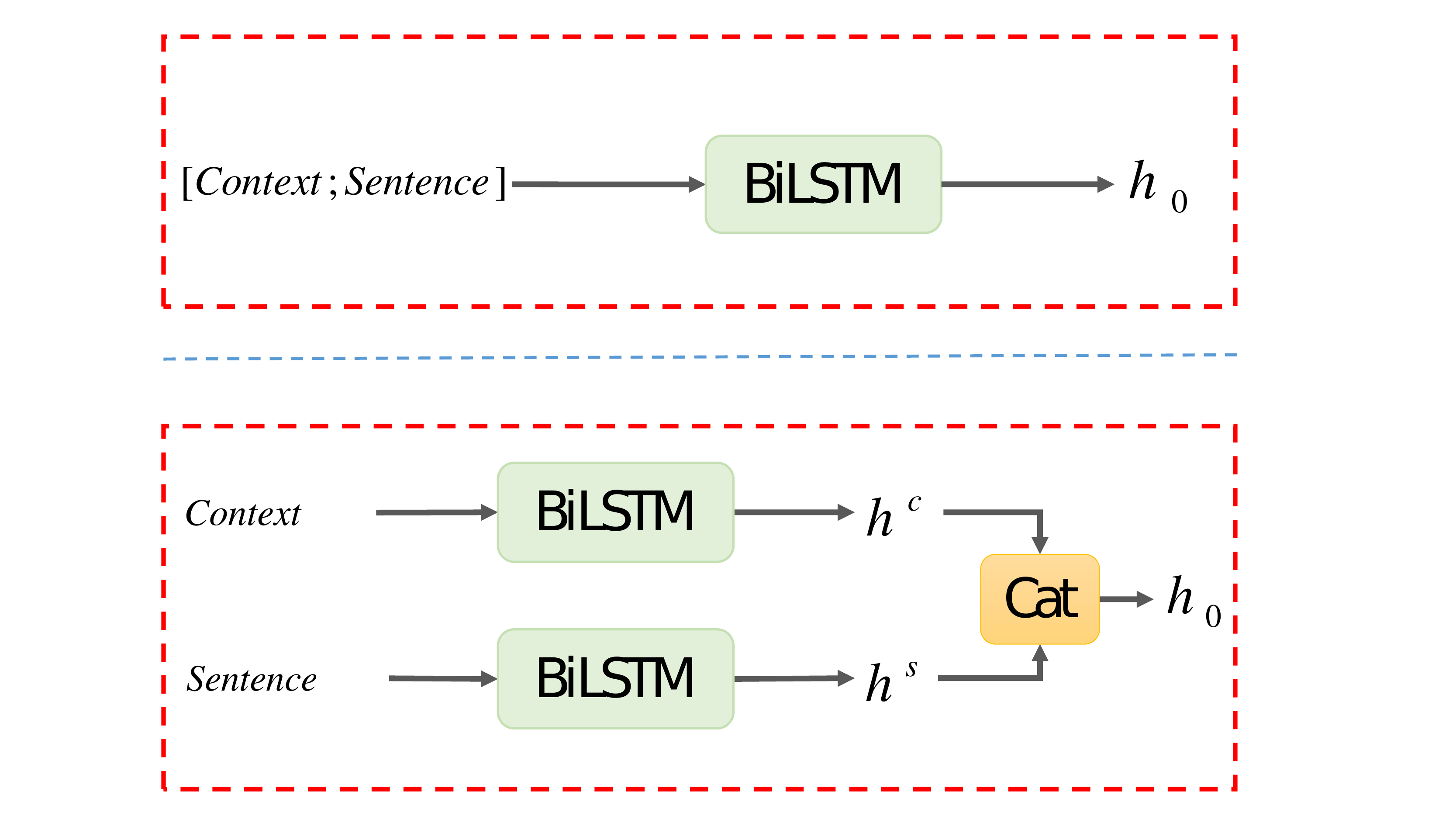}
\label{Encoder}
\end{minipage}
}
\subfigure[Decoder]{
\begin{minipage}[t]{0.45\linewidth}
\centering
\includegraphics[scale=0.15]{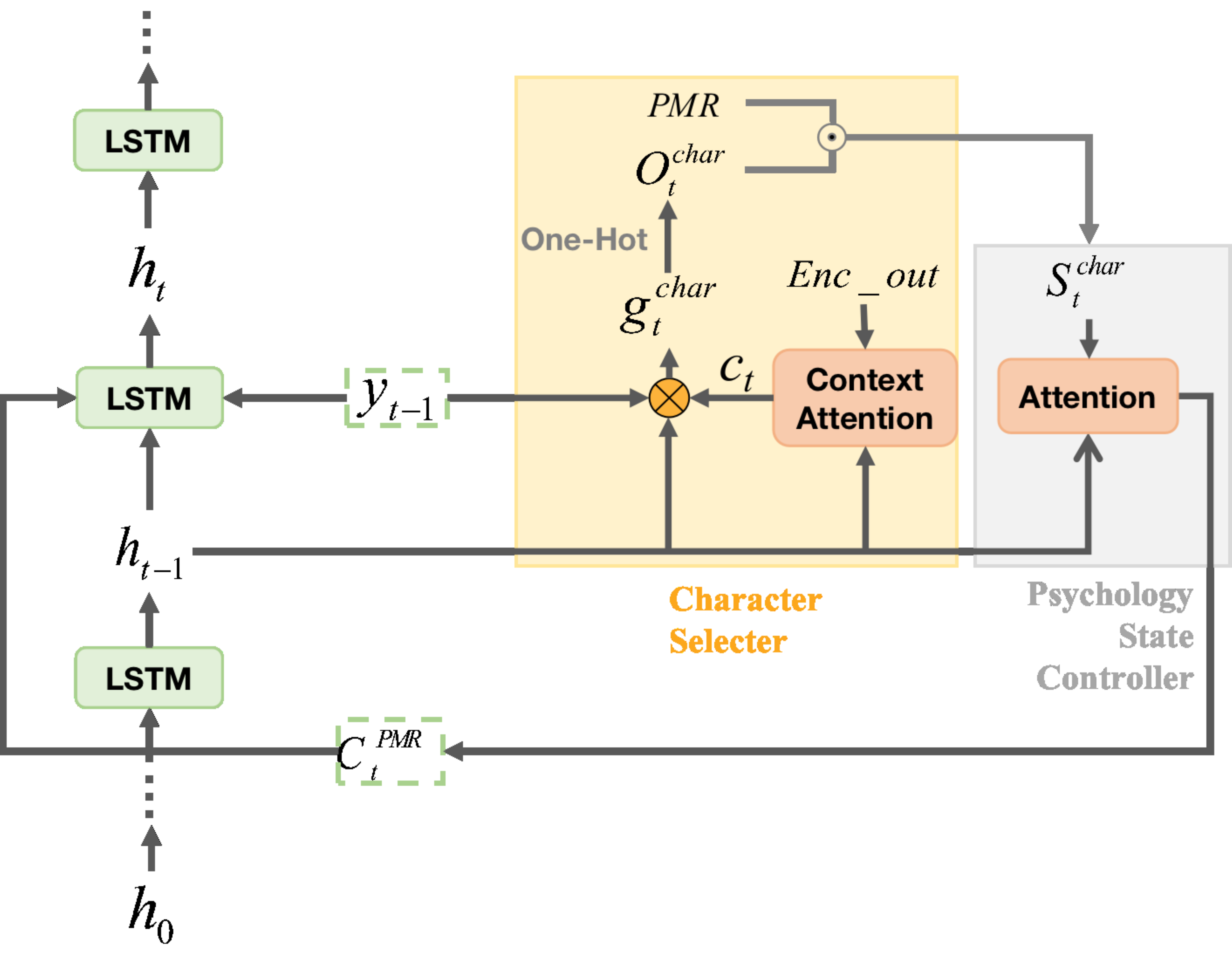}
\label{Decoder}
\end{minipage}
}
\caption{The encoder and decoder modules in the SoCP model. In the Encoder part, we present two methods to encode the text: Context-independent and Context-merge. In the Decoder part, we adopt Character Selector and Psychological State Controller to control text generation.}
\end{figure*}

An overview of our SoCP model is presented in Figure 
\ref{fig2Overview}. Our model aims to generate a multi-character and multi-emotion story. In this task, we design story characters and the psychological state lines of each character as our input in addition to the seed text.
We use a Seq2Seq structure to perform story generation based on the seed text, typically a single input sentence. 
For the purpose of controlling multiple characters and multiple emotions, we design two extra important components called the character selector and the psychological state controller. We first use the character selector to select a person and then use the psychological state controller to attend the three indicators: the Plutchik matrix, Maslow matrix and Reiss matrix. For convenience, in this task, we simply refer to the psychological state as the emotion. 

Formally, there is the input story sentence 
$X = (x{^s_1},...,x{^s_n})$
consists of $n$ words. To make the story coherent, we use the context $C = (x{^c_1},...,x{^c_m})$ into account, which represents previous sentences of the input sentence $X$ and consists of $m$ words. We concatenate C and X as input.
Meanwhile, there is each character’s PMR score matrix:
$S^{pmr} = {(S_1^p, S_1^m, S_1^r),...,(S_j^p, S_j^m, S_j^r)}$
, which represents the three Plutchik, Maslow, and Reiss indicators to describe a character in the story, where $j$ is the number of characters in the story. Our purpose is to generate a sentence $Y = (y{_1},...,y{_k})$ that describes the continuation of the story line. Essentially, our goal is to estimate the conditional probability
$P(Y|[X,C], PMR)$, where $PMR$ is the PMR matrix we designed.
In this work, we assume that 
$S^{pmr}$ 
are given as input for text generation. 

\subsection{PMR matrix}
To add the psychological theories to story generation, we construct a PMR matrix, which is a concatenation of the Plutchik, Maslow, and Reiss matrices.
The three indicators
represent the psychological state of a character in a storyline. 
In the dataset, each indicator has its own score annotated by human annotators, and we use these scores to construct a vector for each indicator. Due to the different number of characters in different stories, we fix the number of characters as the maximal number of characters in all stories. 
A multiple characters score matrix is obtained by concatenating the Plutchik score $S^p$, Maslow score $S^m$, and Reiss score $S^r$. Their dimension is the maximum number of characters multiplied by the number of psychological states.
Then, we randomly initialize a word embedding matrix for three indicators as $V^p$, $V^m$, and $V^r$ respectively, and their dimension is the number of psychological states multiplied by the size of the word embedding.
We multiply the word embedding matrix with multiple characters score matrix and then map them onto a feature space. Subsequently, we obtain the Plutchick, Maslow, and Reiss matrices as follows:
\begin{equation}
    P_{i} = W_p(S^{p}_{i} \times V^{p}_{i}) + b_p,
\end{equation}
\begin{equation}
    M_{i} = W_m(S^{m}_{i} \times V^{m}_{i}) + b_m,
\end{equation}  
\begin{equation}
    R_{i} = W_r(S^{r}_{i} \times V^{r}_{i}) + b_r,
\end{equation}
where $W_p$, $W_m$, $W_r$ are weight matrices for deriving Plutchick, Maslow, and Reiss representations of characters, respectively. $b_p$, $b_m$, $b_r$ are biases and $i$ indicates the $i$-$th$ character. For the convenience of calculation, we can concatenate the Plutchick, Maslow, and Reiss matrices as multiple characters PMR matrix:
\begin{equation}
    PMR_i = W_{pmr}(S^{pmr}_{i}\times V^{pmr}_{i}) + b_{pmr},
\end{equation}
where $W_{pmr}$ and $b_{pmr}$ are learnable weight matrix and bias.

\subsection{Encoder}
\subsection{Encoder}

For the coherence of the generated story, we present two methods to add contextual information into the model. 
We use a normal bidirectional 
\textbf{LSTM} \cite{schuster1997bidirectional} as our encoder module, which is illustrated in Figure \ref{Encoder}.
We do not only encode the input sentence but also consider the contexts, which include the previously generated sentences.

\subsubsection{Context-independent}
The first method, called the context-merger method, concatenate the context and sentence together as input which form the input to the first BiLSTM encoder. Subsequently, we obtain the last hidden state of encoder $h_0$. 
\begin{equation}
    x = (x{^c_0}, x{^c_1}, ..., x{^c_m};x{^s_0}, x{^s_1}, ..., x{^s_n}),
\end{equation}
\begin{equation}
    h_0 = \mathrm{BiLSTM}(x).
\end{equation}

\subsubsection{Context-merge}
In addition, we use two BiLSTM encoders to encode the sentence and context. Then, we concatenate the last hidden state $h{^c}$ of the encoded context and $h{^s}$ of the encoded sentence as text representation that is input to the decoder, which is also the initial hidden state $h{_0}$ of the decoder.  
\begin{equation}
    h{^c} = \mathrm{BiLSTM}^c(x{^c_0}, x{^c_1}, ..., x{^c_m}),
\end{equation}
\begin{equation}
    h{^s} = \mathrm{BiLSTM}^s(x{^s_0}, x{^s_1}, ..., x{^s_n}),
\end{equation}
\begin{equation}
    h{_0} = [h{^c}; h{^s}],
\end{equation}
where the [·;·] represents the concatenation operation.

\subsection{Decoder}
The structure of our decoder is shown in Figure \ref{Decoder}. In the decoder part, we have two auxiliary modules to help generate sentences: the character selector and the psychological state controller.

\subsubsection{Character Selector}
We design a character selector to select which person should be described during the decoding.
At each step $t$, we apply Multi-Layer Perceptron (MLP) to compute a character gate $g_{t}^{char}$ with the input of the word embedded in the previously decoded word $y_{t-1}$, the previous hidden state of the decoder $h_{t-1}$, and the current context vector $c_{t}$. The character gate is defined as follows:
\begin{equation}
    g_{t}^{char} = \mathrm{softmax}(W_{g}[y_{t-1};h_{t-1};c_{t}]),
    \label{char_selector}
\end{equation}
\begin{equation}
    o_{t}^{char} = \mathrm{onehot}(g_{t}^{char}),
\end{equation}
\begin{equation}
    s_{t}^{char} = o_{t}^{char} \times PMR_i,
\end{equation}
where $W_g$ is the weight matrix in character gate. We use softmax activation to select a character of maximum probability. Then, we transform it into a one-hot vector. Finally, $PMR_i$ is multiplied by the one-hot vector to obtain the final psychological state of the selected character.

The above character gate mechanism indicates that the selected character that would be described in step $t$ in the decoding.

\subsubsection{Psychological State Controller}
After the character is selected by the character selector, we use a psychological state controller inspired by \cite{bahdanau2014neural} to control which psychological states and how many we accept. The formulation of PMR context $c_{t}^{PMR}$ at step $t$ is as follows:
\begin{equation}
     e_{t,i} = V_{a}^{T}\mathrm{tanh}(W_{a}s_{t}^{char} + U_{a}h_{t-1}),
\end{equation}
\begin{equation}
    \alpha_{t,i} = \frac{\exp(e_{t,i})}{\sum_{i=1}^{c}\exp(e_{t,i})},
    \label{psy_controller}
\end{equation}
\begin{equation}
    c_{t}^{PMR} = \sum_{i=1}^{c}\alpha_{t,i}s_{t}^{char},
\end{equation}
where $V_a^T$, $W_a$, $U_a$ are learnable parameters, and $c$ is the number of characters.
Through our psychological state controller, the model is effective in learning which psychological state should be attended.

\subsubsection{State Updating}
We use LSTM \cite{hochreiter1997long} with attention as the decoder. We initialize the hidden state of the decoder with the encoded text representation $h_{0}$ from the encoder output. 
The decoder input is the concatenation of the last timestep word embedding vector $y_{t-1}$ and
PMR context $c_t^{PMR}$, which represent adding the psychological state into each character.

We update the LSTM hidden state as follows:

\begin{equation}
    h_{t} = \mathrm{LSTM}([y_{t-1};c_{t}^{PMR};c_t], h_{t-1}),h_{t=0}=h_{0}.
\end{equation}

\subsection{Training}
Our training objective is the negative log likelihood.

\begin{equation}
    L_{NLL} = - \frac{1}{N} \sum_{i=1}^{N}\sum_{t=1}^{T}log P(y_{t}^{(i)}|y_{<t}^{(i)},X^{(i)}, C^{(i)},P^{(i)},M^{(i)},R^{(i)}),
\end{equation}
where $N$ is the total number of the dataset and $T$ is the timestep of the $i$-$th$ output sentence. $X^{(i)}$ represents the $i$-$th$ sentence in the dataset. Similarly, $C^{(i)}$, $P^{(i)}$, $M^{(i)}$ and $R^{(i)}$  represents the $i$-$th$ context, $i$-$th$ Plutchik matrix, $i$-$th$ Maslow matrix and $i$-$th$ Reiss matrix in the dataset respectively. 

We minimum the objective function with the given inputs to generate the proper sentences.

\section{Experiments}
\subsection{Dataset}
\begin{figure}[t]
\centering
\includegraphics[scale=0.3]{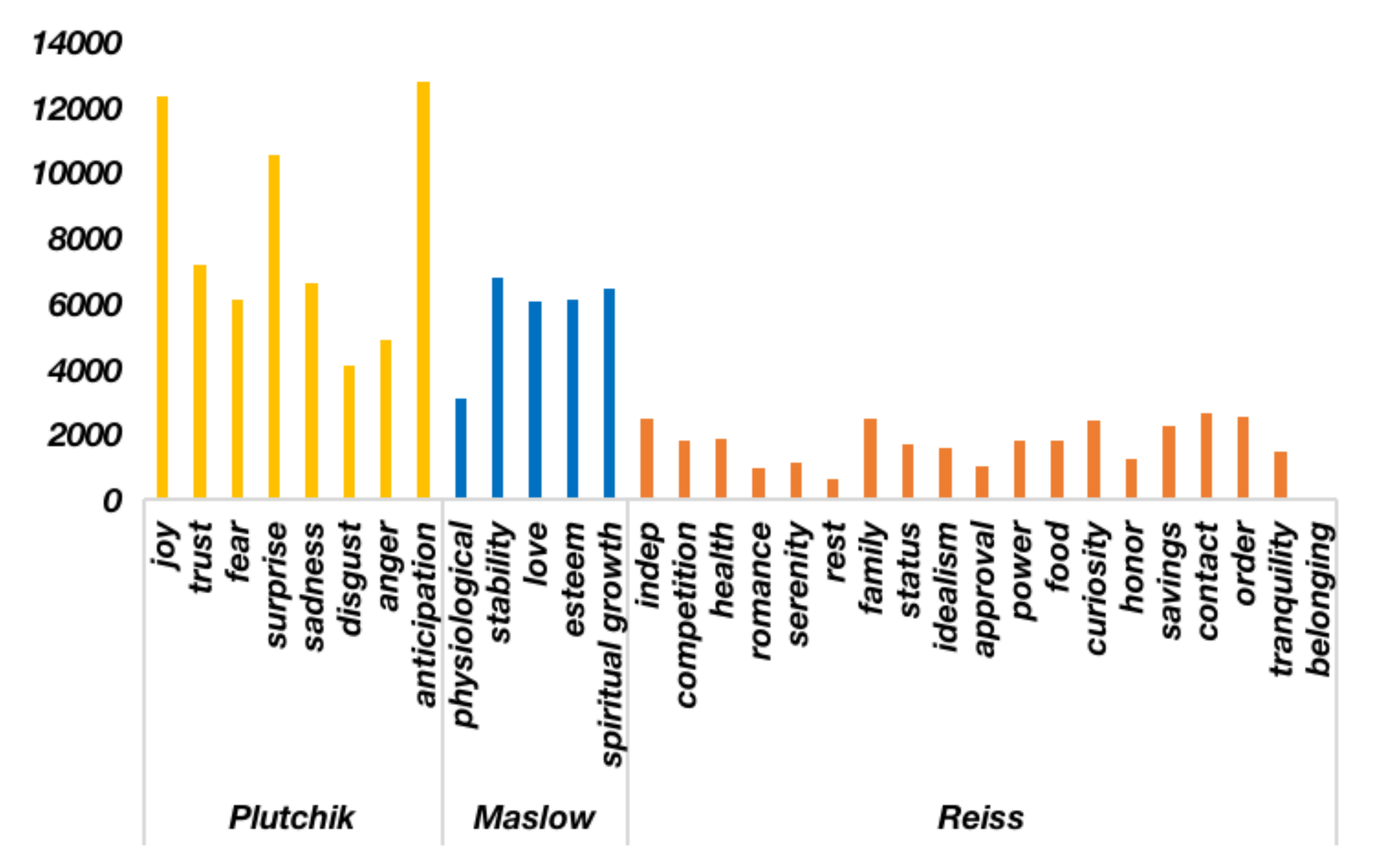}
\caption{The statistic of psychological states. ``Anticipation'' is the most frequent state in Plutchick's wheel of emotion, while ``belonging'' is the least frequent state in the Reiss classification.} 
\label{pmr_statistic}
\end{figure}

We choose a story corpus that has been annotated with psychology theories \cite{rashkin-etal-2018-modeling}
, and consists of 4k five-sentence stories. Therefore we have 20k sentences. 
Each sentence is annotated with characters and three psychological theory indicators.
In psychological theory, Plutchik has 8 fine-grained categories, Maslow has five coarse-grained categories, and Reiss has 19 fine-grained categories. Figure \ref{pmr_statistic} shows the statistics of the psychological states.
It shows that Plutchik is the most frequently annotated psychological state compared to Maslow and Reiss. In particular, among the annotated Plutchik states, "joy" and "participation" are the most frequent states. 
Specifically, the Reiss categories can be considered as subcategories of the Maslow categories. 
We have processed the dataset by different methods for the \textsl{Plutchik}, \textsl{Maslow} and \textsl{Reiss} indicators in the following way. 
Since the original data are annotated by three workers and since different workers have their own viewpoint, we add them together and obtain only the score that is marked by more than two workers for the Plutchik indicator and then we normalize it. The Maslow and Reiss are represented as one-hot vector.
We randomly split the data, and use 80\% of the data as training set and 20\% as test set.
However, due to the lack of enough stories, we adopt a classifier proposed in \cite{paul-frank-2019-ranking} to label the psychological states of characters in the unannotated stories and use them as augmented data.
By including the augmented data, we have approximately 15k stories in total. 
In the test phase, we use the beginning sentences, contexts, and the normalized scores of psychological states as inputs. 
The character number statistics are shown in Table \ref{statistic}. The upper limit of the number of characters is six, and the number of characters in a story is mostly distributed in the range of 1-3. Thus, we fix the number of characters to 3, which covers most of the training data.

\begin{table}[t]
\centering
\caption{The statistic of the number of characters in sentences.}
\begin{tabular}{cc}
\hline
\textbf{Character number} & \textbf{Sentence number} \\ 
\hline 
1 & 11008 \\ 
2 & 6547 \\
3 & 1254 \\
4 & 164 \\
5 & 36 \\
6 & 1 \\
\hline
\end{tabular}
\label{statistic}
\end{table}

\begin{table*}[htbp]
\caption{Automatic evaluations of the proposed model and the baseline models. The last line with * indicates that the results are improved using data augmentation.}
\centering
\resizebox{\textwidth}{25mm}{
\begin{tabular}{c|cccc|ccc|c|c}
\hline
\multirow{2}{*}{\textbf{Model}} & \multicolumn{4}{c|}{\textbf{BLEU}} & \multicolumn{3}{c|}{\textbf{ROUGE}} & \multirow{2}{*}{\textbf{Meteor}} & \multirow{2}{*}{\textbf{ACER}} \\ \cline{2-8}
 & \textbf{B-1} & \textbf{B-2} & \textbf{B-3} & \textbf{B-4} & \textbf{R-1} & \textbf{R-2} & \textbf{R-l} &  & \\ \hline
Seq2Seq & 0.202 & 0.049 & 0.016 & 0.007 & 0.122 & 0.014 & 0.117 & 0.062 & 0.654 \\ 
Seq2Seq+context-merger & 0.208 & 0.049 & 0.016 & 0.008 & 0.134 & 0.012 & 0.126 & 0.065 & 0.674\\ 
Seq2Seq+context-independent & 0.223 & 0.055 & 0.018 & 0.009 & 0.147 & 0.016 & 0.138 & 0.068 & 0.684 \\ 
\hline
Transformer+context & 0.207 & 0.053 & 0.020 & 0.008 & 0.139 & 0.012 & 0.128 & 0.069 & 0.747 \\
Inc-S2S+context & 0.224 & 0.053 & 0.017 & 0.006 & 0.151 & 0.013 & 0.141 & 0.067 & 0.825 \\ 
\hline
(Ours)SoCP+context-merger & 0.216 & 0.056 & 0.021 & 0.010 & 0.144 & 0.016 & 0.136 & 0.067 & 0.886 \\ 
(Ours)SoCP+context-independent & 0.232 & 0.062 & 0.025 & 0.011 & 0.161 & 0.018 & 0.151 & 0.072 & 0.879 \\
(Ours)SoCP+context-independent* & \textbf{0.241} & \textbf{0.070} & \textbf{0.029} & \textbf{0.014} & \textbf{0.167} & \textbf{0.022} & \textbf{0.158} & \textbf{0.076} & \textbf{0.893} \\
\hline
\end{tabular}
\label{Automatic}
}
\end{table*}

\subsection{Baselines}
To evaluate the effectiveness of the SoCP framework, we compare our methods against representative baselines as follows:

\textbf{Seq2Seq} is a generation model that has good performance in the translation task and other generation tasks. It is also the foundation model of our SoCP. Therefore, we use the Seq2Seq baseline as a comparison metric to prove whether our model can improve the traditional Seq2Seq and whether it exerts a generation effect on fluency.

\textbf{Inc-S2S}, mentioned in \cite{yao2019plan}, 
denotes the incremental sentence-to-sentence generation baseline. Our experiment differs from this implementation in that we do not have the story title, we appoint the psychological states to the model, and we generate the next sentence from the given first sentence of the story and previously generated sentences. We also use the Seq2Seq model, paying attention to construct the Inc-S2S baseline. In this way, we can prove whether our character selector and psychological state controller work effectively.

\textbf{Transformer} architecture \cite{10.5555/3295222.3295349} has attracted the attention of many scholars in recent years. It greatly improves the state of the art in natural language processing tasks by training on a huge amount of data with large-scale language models (LMs), such as BERT \cite{devlin-etal-2019-bert}, GPT-2  \cite{radford2019language}, and Transformer-xl \cite{dai-etal-2019-transformer}.

\textbf{GPT-2} \cite{radford2019language} is well known in natural language processing models in recent years because of its huge number of parameters, and it also composed of a Transformer structure. Many works have proven that GPT-2 can  generate a reasonable and personalized story. Therefore, we use GPT-2, a representative of story generation models, as a benchmark to our model.

\subsection{Experimental Settings}
Following the discussion above, we fix the number of characters to 3 and use ‘none’ to represent the character if the number of characters is smaller than 3.
300-dimensional pre-trained glove vectors are used as the word embeddings.
The encoder is designed as a 2-layer bidirectional LSTM with a hidden size of 256 and the decoder as a 1-layer LSTM with a hidden size of 256.
The PMR matrix is used to from a high-dimensional to a low-dimensional vector with dimension 256. 
The batch size is 8, and the dropout \cite{srivastava2014dropout} is set as 0.2. We use the Adam optimizer \cite{Adam} with an initial learning rate of 0.0003. 

\begin{figure*}[htbp]
\centering
\subfigure[attention map 1]{
\begin{minipage}[t]{0.45\linewidth}
\centering
\includegraphics[scale=0.25]{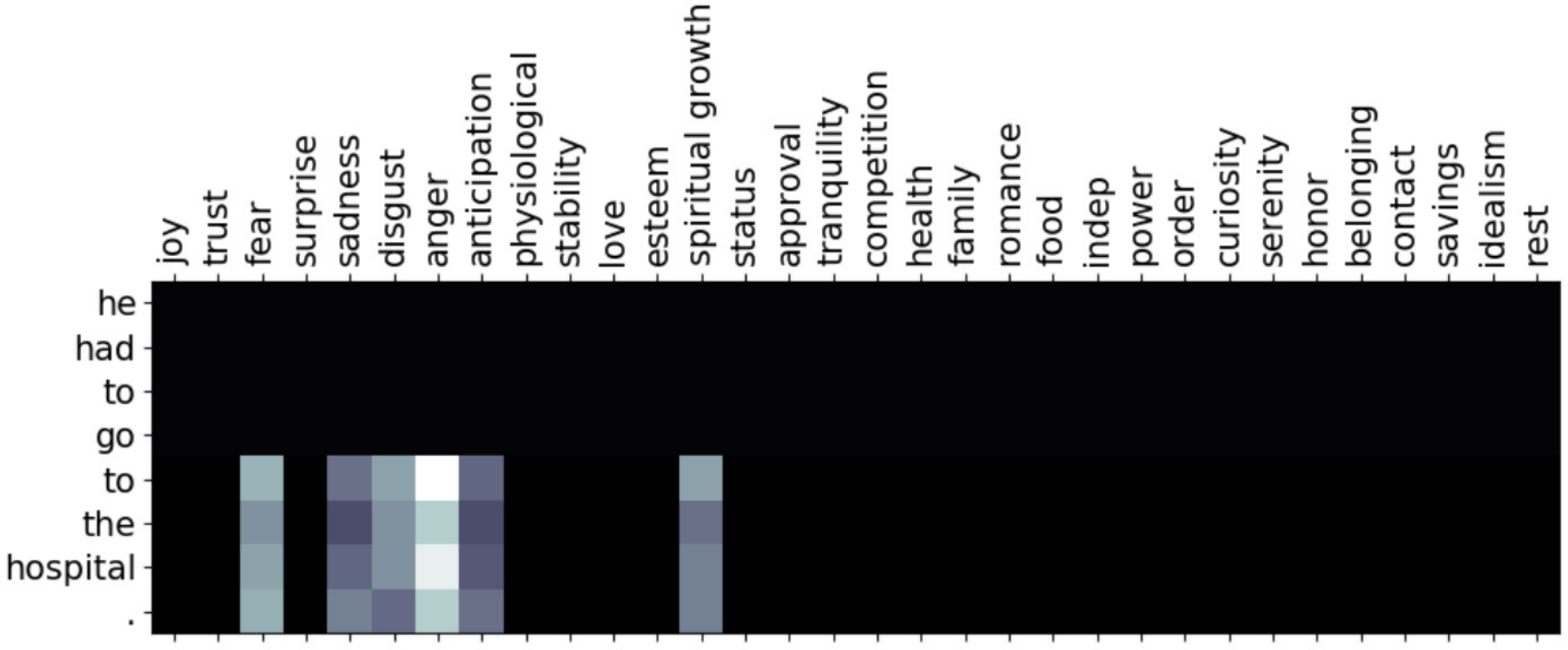} 
\label{map1}
\end{minipage}
}
\subfigure[attention map 2]{
\begin{minipage}[t]{0.45\linewidth}
\centering
\includegraphics[scale=0.25]{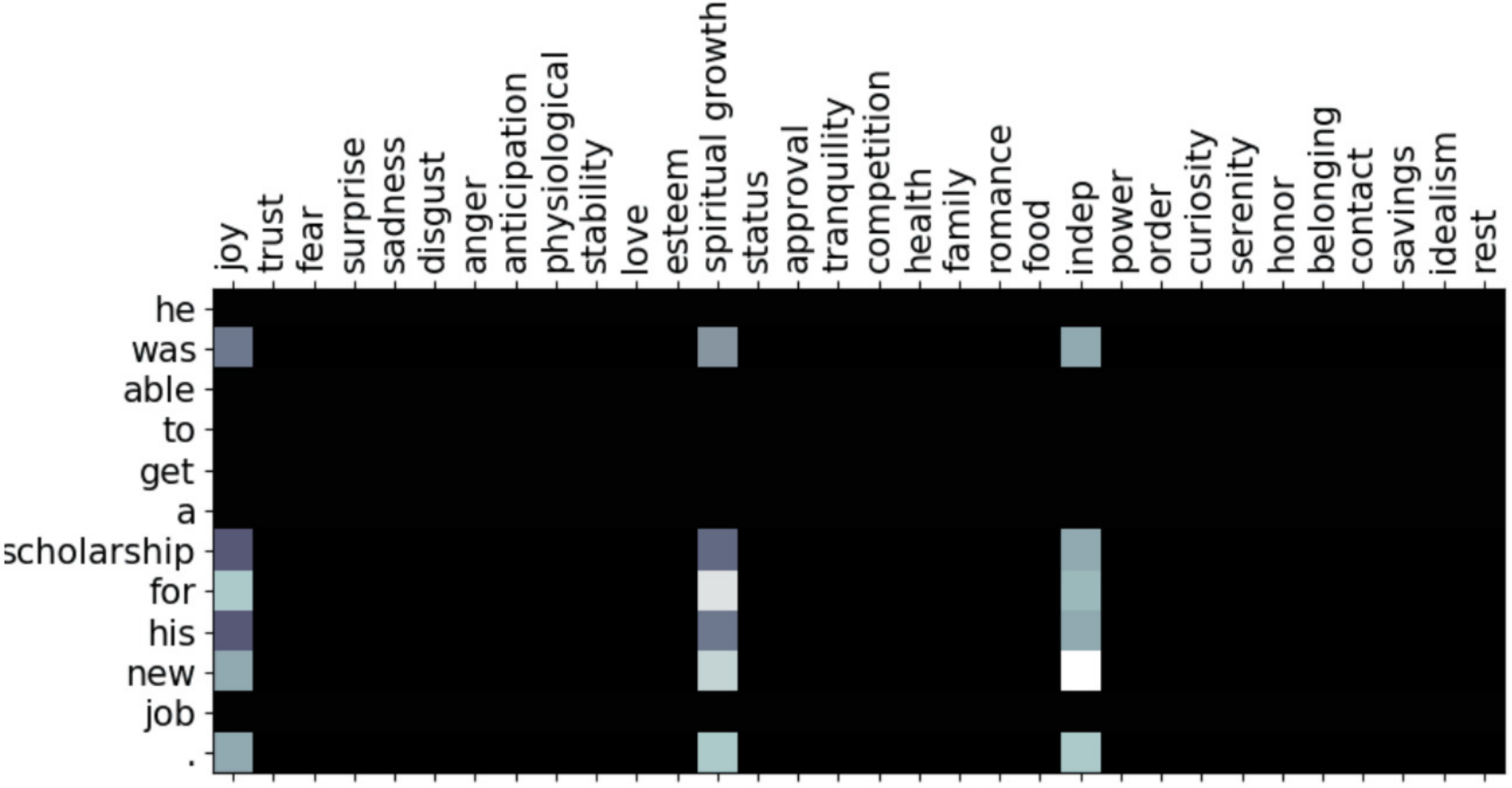}
\label{map2}
\end{minipage}
}
\subfigure[attention map 3]{
\begin{minipage}[t]{0.45\linewidth}
\centering
\includegraphics[scale=0.25]{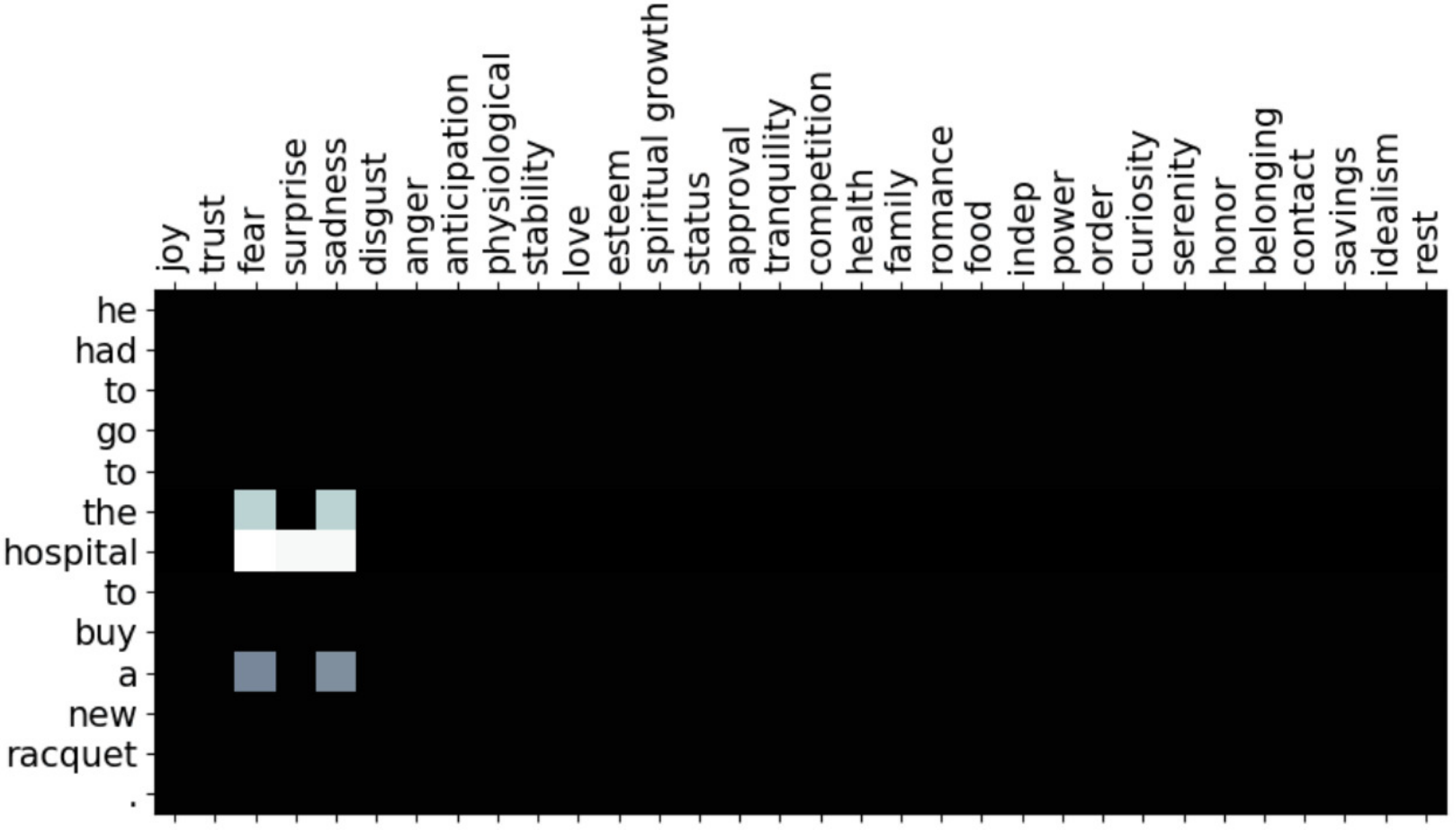} 
\label{map3}
\end{minipage}
}
\subfigure[attention map 4]{
\begin{minipage}[t]{0.45\linewidth}
\centering
\includegraphics[scale=0.25]{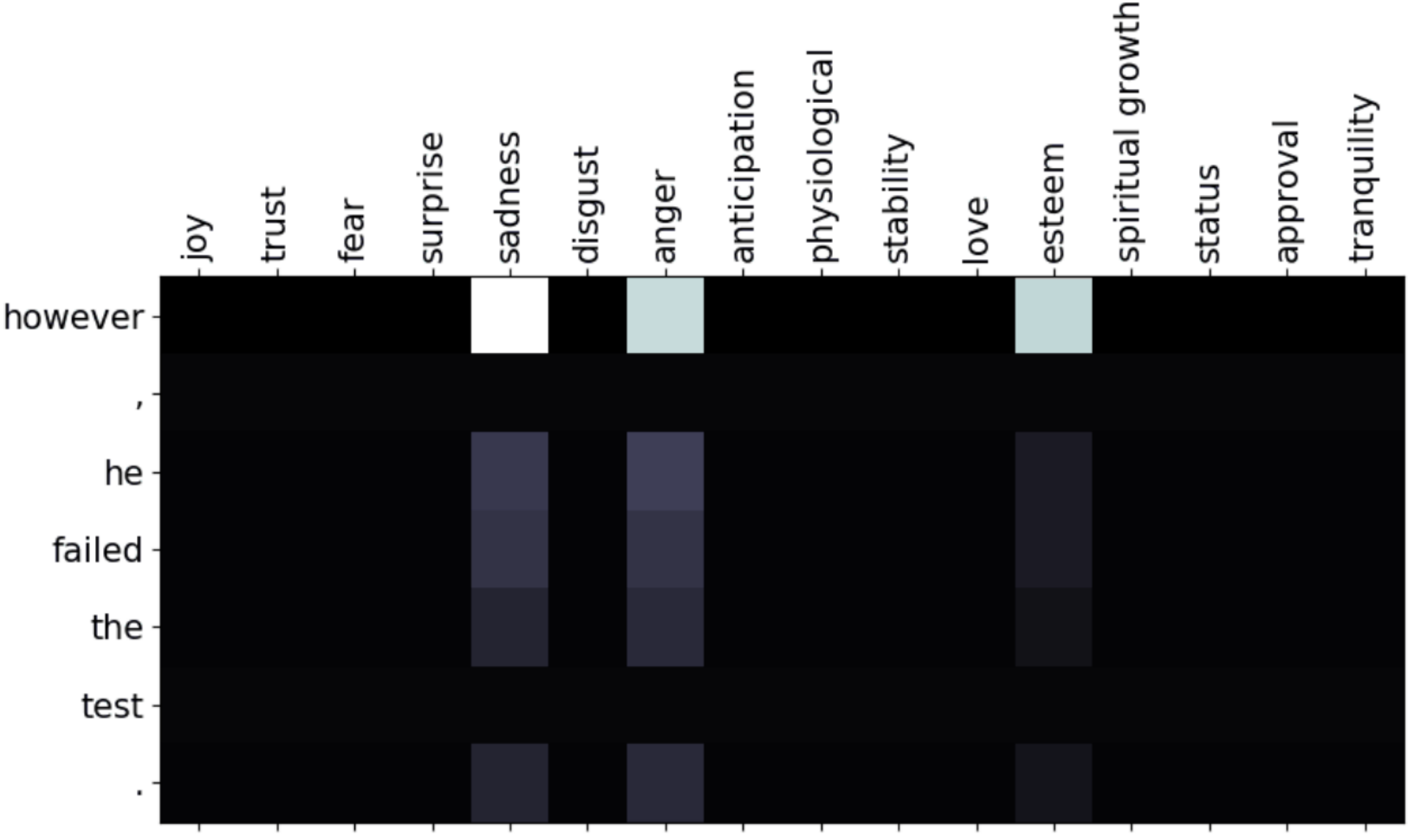}
\label{map4}
\end{minipage}
}
\centering
\caption{Visualization  of  our  psychological state controller. The  row  in  the  figure  represents  the  fine-grained emotion and motivation of the three psychological states, and the column is the generated sentence. The brighter the grid in the graph is, the more there is attention between rows and columns.}
\label{map}
\end{figure*}

\subsection{Evaluation Metrics}
\textbf{BLEU}: For each story in the testset, we take the input sentence $x$ and the human-annotated processed PMR score $S^{pmr}$ as input. The corresponding label is $\hat{y}$.
Then, we calculate the BLEU \cite{papineni-etal-2002-bleu} score of the generated $y$ and $\hat{y}$ as the overall quality of the generated story.

\textbf{ROUGE}: Similar to BLEU, ROUGE is also a common evaluation indicator in machine translation and text summarization. It is used to measure the similarity between automatically generated and reference results.

\textbf{Meteor}: Meteor considers word shape and is also an important evaluation indicator of fluency.

\textbf{ACER} (Accuracy of Controlling Emotions of Roles): We use BiLSTM to pretrain a classifier to classify the psychological state of each character in the generated sentences. It indicates our model’s capability of expressing emotion. Because of the complexity of psychological state, we design a classifier only for the Plutchik indicator.
We concatenate the name of a story character $Char$ and the sentence $X$ as the input. In this way, many training pairs in identical sentences with different outputs can be obtained because different characters have different psychological states in a sentence. Through BiLSTM, we can obtain the compressed vector $h^{clf}$, and then, we use two fully connected layers to map it onto the output size.

\begin{equation}
    h^{clf} = \mathrm{BiLSTM}([Char;X]),
\end{equation}
\begin{equation}
    p = W_p(\mathrm{ReLU}(W_o(h^{clf}) + b_o))+b_p,
\end{equation}
where $W_p$, $W_o$, $b_o$, and $b_p$ are learnable parameters.

\begin{table*}[htbp]
\caption{Examples generated by the proposed model and the baseline models.}
\resizebox{\textwidth}{!}{
\begin{tabular}{|c|l|l|}
\hline
\multirow{5}{*}{\textbf{Input}} & \textbf{Jane bought a new necklace.} & \textbf{The man next door seemed to be losing his mind.} \\ \cline{2-3} 
 & \multirow{4}{*}{\textbf{\begin{tabular}[c]{@{}l@{}}Characters: Jane, Friend\\ Plutchik: joy-joy-joy-joy-joy, none-none-joy-none-none\\ Maslow: esteem, love\\ Reiss:approval\end{tabular}}} & \multirow{4}{*}{\textbf{\begin{tabular}[c]{@{}l@{}}Characters: Man\\ Plutchik: fear-disgust-surprise-surprise-joy\\ Maslow: spiritual growth\\ Reiss: status\end{tabular}}} \\
 &  &  \\
 &  &  \\
 &  &  \\ \hline
\multirow{5}{*}{\textbf{\begin{tabular}[c]{@{}c@{}} Seq2Seq
\end{tabular}}} & \textbf{Jane bought a new necklace.} & \textbf{The man next door seemed to be losing his mind.} \\ \cline{2-3} 
 & \multirow{4}{*}{\begin{tabular}[c]{@{}l@{}}She was running it.\\ she was amazed it.
\\  She tried it.\\ She was thankful jane it.\end{tabular}} & \multirow{4}{*}{\begin{tabular}[c]{@{}l@{}}He was very happy to have a new job.\\ He was not sure what he was going.\\ He was not sure what he was going.\\ He was not sure to go.\end{tabular}} \\
 &  &  \\
 &  &  \\
 &  &  \\ \hline
\multirow{5}{*}{\textbf{Inc-S2S}} & \textbf{Jane bought a new necklace.} & \textbf{The man next door seemed to be losing his mind.} \\ \cline{2-3} 
 & \multirow{4}{*}{\begin{tabular}[c]{@{}l@{}}She was not sure what she wanted to go to the store.\\ She was very happy.\\ She was very happy.\\ She was so happy that she had a good time.\end{tabular}} & \multirow{4}{*}{\begin{tabular}[c]{@{}l@{}}He asked his credit hour he did not know how.\\ He asked his parents would help him some directions.\\ He tied the games and tied miles.\\ He was sad he realized he realized that his card.\end{tabular}} \\
 &  &  \\
 &  &  \\
 &  &  \\ \hline
\multirow{5}{*}{\textbf{Transformer}} & \textbf{Jane bought a new necklace.} & \textbf{The man next door seemed to be losing his mind.} \\ \cline{2-3} 
 & \multirow{4}{*}{\begin{tabular}[c]{@{}l@{}}She decided to do something about it.\\ She did not have any money.\\ She decided to save money.\\ She decided to save up.\end{tabular}} & \multirow{4}{*}{\begin{tabular}[c]{@{}l@{}}He yelled at him.\\ He had to go to the hospital.\\ He asked his girlfriend.\\ She said yes right away.\end{tabular}} \\
 &  &  \\
 &  &  \\
 &  &  \\ \hline
 \multirow{5}{*}{\textbf{GPT-2}} & \textbf{Jane bought a new necklace.} & \textbf{The man next door seemed to be losing his mind.} \\ \cline{2-3} 
 & \multirow{4}{*}{\begin{tabular}[c]{@{}l@{}}"I'm not sure if I'm going to be able to wear it anymore," she said. \\ "I'm not sure if I'm going to be able to wear it anymore."\\ She said she's been looking for a way to get rid of the necklace.\\ "I'm not sure if I'm going to be able to wear it anymore," she said. \end{tabular}} & \multirow{4}{*}{\begin{tabular}[c]{@{}l@{}}"I'm sorry, but I'm not going to be able to go to the hospital," he said. \\ "I'm not going to be able to go to the hospital."\\ The man next door was also in shock.\\ "I'm not going to be able to go to the hospital," he said. \end{tabular}} \\
 &  &  \\
 &  &  \\
 &  &  \\ \hline
\multirow{5}{*}{\textbf{SoCP}} & \textbf{Jane bought a new necklace.} & \textbf{The man next door seemed to be losing his mind.} \\ 
\cline{2-3} 
& \multirow{4}{*}{\begin{tabular}[c]{@{}l@{}}She was excited to get a job.\\ Her friend was on the same.\\ She was very happy.\\ She was happy to the best game.\end{tabular}} & \multirow{4}{*}{\begin{tabular}[c]{@{}l@{}}He was very angry and did not want to do.\\ He was a detective man about the problem.\\ The man was not as the man was not the wrong.\\ He was able to beat the ball and he was shot.\end{tabular}} \\
 &  &  \\
 &  &  \\ 
 &  &  \\ \hline
\end{tabular}}
\label{examples1}
\end{table*}

\begin{table*}[htbp]
\caption{Examples of controllable generated stories}
\small
\begin{tabular}{llll}
\hline
\multicolumn{2}{|l|}{\textbf{Input sentence}} & \multicolumn{2}{l|}{\textbf{Jane bought a new necklace.}} \\ \hline
\multicolumn{2}{|l|}{\textbf{Character}} & \multicolumn{2}{l|}{Jane} \\ \hline
\multicolumn{2}{|l|}{\textbf{Maslow}} & \multicolumn{2}{l|}{esteem, love} \\
\multicolumn{2}{|l|}{\textbf{Reiss}} & \multicolumn{2}{l|}{approval} \\ \hline
\multicolumn{4}{|c|}{\textbf{Generated stories with the same Plutchik indicator "joy" under different scores}} \\ \hline
\multicolumn{1}{|c|}{\textbf{joy=1}} & \multicolumn{2}{c}{\textbf{joy=0.5}} & \multicolumn{1}{|c|}{\textbf{joy=0}} \\ \hline
\multicolumn{1}{|l|}{\textbf{Jane bought a new necklace.}} & \multicolumn{2}{l|}{\textbf{Jane bought a new necklace.}} & \multicolumn{1}{l|}{\textbf{Jane bought a new necklace.}} \\
\multicolumn{1}{|l|}{She was a good amused.} & \multicolumn{2}{l|}{She was a really excited.} & \multicolumn{1}{l|}{She needed to go to a different game for a month.} \\
\multicolumn{1}{|l|}{She went to school.} & \multicolumn{2}{l|}{She decided to have a swim.} & \multicolumn{1}{l|}{She was nervous about her feet.} \\
\multicolumn{1}{|l|}{She was a happy.} & \multicolumn{2}{l|}{She did not think it was on it.} & \multicolumn{1}{l|}{She was so tired, they got there for the store.} \\
\multicolumn{1}{|l|}{She was proud of her career.} & \multicolumn{2}{l|}{It was the best she ever.} & \multicolumn{1}{l|}{Her mom was not happy.} \\ \hline

\multicolumn{4}{|c|}{\textbf{Generated Stories with Different Plutchik Indicators}} \\ \hline
\multicolumn{1}{|c|}{\textbf{surprise=1}} & \multicolumn{2}{c}{\textbf{fear=1}} & \multicolumn{1}{|c|}{\textbf{anger=1}} \\ \hline
\multicolumn{1}{|l|}{\textbf{Jane bought a new necklace.}} & \multicolumn{2}{l|}{\textbf{Jane bought a new necklace.}} & \multicolumn{1}{l|}{\textbf{Jane bought a new necklace.}} \\
\multicolumn{1}{|l|}{She was surprised to see what she had been.} & \multicolumn{2}{l|}{She had no idea of her own.} & \multicolumn{1}{l|}{She was angry at a slow game.} \\
\multicolumn{1}{|l|}{The day was a huge.} & \multicolumn{2}{l|}{She was afraid of heights.} & \multicolumn{1}{l|}{The kid removed the flame on the drums.} \\
\multicolumn{1}{|l|}{I did not know how to stop.} & \multicolumn{2}{l|}{She had to swim on stage.} &  \multicolumn{1}{l|}{it was too late.} \\
\multicolumn{1}{|l|}{I was shocked when i looked at the door.} & \multicolumn{2}{l|}{She had to be scared to be scared.} & \multicolumn{1}{l|}{It turns out a knife to be an important.} \\ \hline

\multicolumn{4}{|c|}{\textbf{Generated Stories with Multiple Plutchik Indicators}} \\ \hline
\multicolumn{1}{|c|}{\textbf{trust=1, joy=0.5}} & \multicolumn{2}{c}{\textbf{sadness=0.4, anticipation=0.8}} & \multicolumn{1}{|c|}{\textbf{fear=1, disgust=0.5}} \\ \hline
\multicolumn{1}{|l|}{\textbf{Jane bought a new necklace.}} & \multicolumn{2}{l|}{\textbf{Jane bought a new necklace.}} & \multicolumn{1}{l|}{\textbf{Jane bought a new necklace.}} \\
\multicolumn{1}{|l|}{She was a book.} & \multicolumn{2}{l|}{ She had never been very busy.} & \multicolumn{1}{l|}{She had a tooth coming in a dark.} \\
\multicolumn{1}{|l|}{She asked her mom to borrow some advice.} & \multicolumn{2}{l|}{Her husband had crashed by.} & \multicolumn{1}{l|}{She was nervous, but she was going to be late.} \\
\multicolumn{1}{|l|}{Her boyfriend bought her a chinese outfit.} & \multicolumn{2}{l|}{Her mother had crashed.} & \multicolumn{1}{l|}{She was afraid of heights.} \\
\multicolumn{1}{|l|}{Her boyfriend promised her to teach her hair.} & \multicolumn{2}{l|}{her mother had to be an prepared.} & \multicolumn{1}{l|}{She was afraid of the back, but she felt very scared.} \\ \hline
\label{examples2}
\end{tabular}
\end{table*}

\begin{table*}[htbp]
\caption{Examples of controllable generated stories}
\resizebox{\textwidth}{!}{
\begin{tabular}{llll}
\hline
\multicolumn{2}{|l|}{\textbf{Input sentence}} & \multicolumn{2}{l|}{\textbf{Tiffany was a defensive soccer player who had never scored a goal.}} \\ \hline
\multicolumn{2}{|l|}{\textbf{Character}} & \multicolumn{2}{l|}{Tiffany} \\ \hline
\multicolumn{2}{|l|}{\textbf{Maslow}} & \multicolumn{2}{l|}{esteem} \\
\multicolumn{2}{|l|}{\textbf{Reiss}} & \multicolumn{2}{l|}{competition} \\ \hline
\multicolumn{4}{|c|}{\textbf{Generated stories with the same Plutchik indicator "joy" under different scores}} \\ \hline
\multicolumn{1}{|c|}{\textbf{joy=1}} & \multicolumn{2}{c}{\textbf{joy=0.5}} & \multicolumn{1}{|c|}{\textbf{joy=0}} \\ \hline
\multicolumn{1}{|l|}{\textbf{Tiffany was a defensive soccer player who ...}} & \multicolumn{2}{l|}{\textbf{Tiffany was a defensive soccer player who ...}} & \multicolumn{1}{l|}{\textbf{Tiffany was a defensive soccer player who ...}} \\
\multicolumn{1}{|l|}{She was so excited and decided to go see the road.} & \multicolumn{2}{l|}{She was a total with the sidewalk.} & \multicolumn{1}{l|}{She was not a personal when he heard a running to the sink.} \\
\multicolumn{1}{|l|}{The resulting were very little and she visited.} & \multicolumn{2}{l|}{She was a very hard.} & \multicolumn{1}{l|}{She was a bit scared, but the water was not working.} \\
\multicolumn{1}{|l|}{It was a great success.} & \multicolumn{2}{l|}{She was able to get things from work.} & \multicolumn{1}{l|}{The manager exchanged the product , but was hard.} \\
\multicolumn{1}{|l|}{I was glad that she had learned to pay a new shoes.} & \multicolumn{2}{l|}{She was glad he had not clean it!} & \multicolumn{1}{l|}{The manager had to move, and they were able to meet.} \\ \hline

\multicolumn{4}{|c|}{\textbf{Generated Stories with Different Plutchik Indicators}} \\ \hline
\multicolumn{1}{|c|}{\textbf{surprise=1}} & \multicolumn{2}{c}{\textbf{fear=1}} & \multicolumn{1}{|c|}{\textbf{anger=1}} \\ \hline
\multicolumn{1}{|l|}{\textbf{Tiffany was a defensive soccer player who ...}} & \multicolumn{2}{l|}{\textbf{Tiffany was a defensive soccer player who ...}} & \multicolumn{1}{l|}{\textbf{Tiffany was a defensive soccer player who ...}} \\
\multicolumn{1}{|l|}{She was surprised to see what he was going to go.} & \multicolumn{2}{l|}{She had been afraid of them since there was no signal.} & \multicolumn{1}{l|}{She tried to use the bigger, but the computer got very hot.} \\
\multicolumn{1}{|l|}{She was halfway there, and the power was coming up.} & \multicolumn{2}{l|}{She tried to be careful when she was there to be at last.} & \multicolumn{1}{l|}{The computer broke down and the bread broke down.} \\
\multicolumn{1}{|l|}{The soup was a bit, but it almost!} & \multicolumn{2}{l|}{she was so nervous, because she had to be stuck in her bed.} &  \multicolumn{1}{l|}{It was also a long time for the driver.} \\
\multicolumn{1}{|l|}{She was shocked to see her daughter.} & \multicolumn{2}{l|}{she was very scared, but the distance went through the door.} & \multicolumn{1}{l|}{It was the last time ever ever could not play football.} \\ \hline

\multicolumn{4}{|c|}{\textbf{Generated Stories with Multiple Plutchik Indicators}} \\ \hline
\multicolumn{1}{|c|}{\textbf{trust=1, joy=0.5}} & \multicolumn{2}{c}{\textbf{sadness=0.4, anticipation=0.8}} & \multicolumn{1}{|c|}{\textbf{fear=1, disgust=0.5}} \\ \hline
\multicolumn{1}{|l|}{\textbf{Tiffany was a defensive soccer player who ...}} & \multicolumn{2}{l|}{\textbf{Tiffany was a defensive soccer player who ...}} & \multicolumn{1}{l|}{\textbf{Tiffany was a defensive soccer player who ...}} \\
\multicolumn{1}{|l|}{She was a simple with the paper.} & \multicolumn{2}{l|}{She had not been very busy at all, and watched a year.} & \multicolumn{1}{l|}{She had been afraid of food since there was no news.} \\
\multicolumn{1}{|l|}{She decided to make a budget.} & \multicolumn{2}{l|}{She asked for a long time to play for 10 minutes.} & \multicolumn{1}{l|}{She tried to be careful when she was all over town.} \\
\multicolumn{1}{|l|}{She was very well.} & \multicolumn{2}{l|}{The hiring on the next day, She was nervous for almost.} & \multicolumn{1}{l|}{She was afraid that she would be an hour for there.} \\
\multicolumn{1}{|l|}{She was able to borrow a new tiesa.} & \multicolumn{2}{l|}{The irony was a janitor that was looking at the end.
} & \multicolumn{1}{l|}{She was very scared and scared of the safety!} \\ \hline
\label{examples3}
\end{tabular}
}
\end{table*}

\subsection{Evaluation}
To demonstrate the effectiveness of our components,
we compare our model with the Seq2Seq structure, the Inc-S2S and Transformer baselines in the BLEU, ROUGE, and Meteor automatic evaluation metrics and our designed metric ACER, which indicates the accuracy with which the psychological state of the generated sentence is consistent with our previous set. As shown in Table \ref{Automatic}, every criteria score in our model is the maximum, which indicates the effectiveness of our designed modules, and the generated sentences are smooth.

For BLEU, ROUGE and Meteor, we see that Seq2Seq has better performance than the Transformer structure. The reason may be that most of the sentences in the data are short text, and it is more suitable to use Seq2Seq. Additionally, it is better to use the context-independent method than the context-merge method with Seq2Seq, which is mentioned in the encoder part. Of all the models, our proposed model has the best performance.

For ACER, which we design to test whether the emotion of the sentences we generated is consistent with our previous settings.
Inc-S2S is better than other models but worse than our model, which proves that our proposed model with our designed component works effectively.

\subsection{Interpretability}
In this part, we study the learned attention distributions of characters' psychological states between generated sentences and three psychological theories, i.e., of Plutchik, Maslow and Reiss.
The model provides interpretability in two ways: by selecting a character based on the character selector using Eq.\ref{char_selector} and by choosing the psychological state based on the psychological state controller using Eq.\ref{psy_controller}, as shown in Figure \ref{map}. The lighter the square at the intersection between two words, the more attention the model pays to psychological states when generating the next word. 
\subsubsection{The Effect of the Character Selector}
As previously mentioned, we proposed the character selector module to select a character that we would like to describe in the current timestep. Visualization of the attention maps provides evidence of the ability of the model to capture relevant psychological states that belong to each person. 
Each word in the sentence corresponds to squares of different colors, indicating that our component can automatically read the psychological state of multiple characters we input and can automatically choose which character's psychological state to accept.
The black square indicates that the character selector does not obtain any responsible psychological state because not all words, such as ‘a’ and ‘the’, necessarily express the emotion.
In summary, our model can be controlled by our designed component to accept external information, similar to our psychological state.

\subsubsection{The Effect of the Psychological State Controller}
Figure \ref{map} shows examples of which psychological state and how much can be attended by the psychological state controller. In the examples shown, the model correctly picks the indicators from the input psychological state.
As shown in Figure \ref{map}, the first attention map has the focus on the Plutchik's emotions, such as ``fear'' and ``anger'', while the second attention map attends to the indicators of Maslow and Reiss, such as ``spiritual growth'' and ``indep''.  
In the third attention map, the word "hospital" is associated with the Plutchik indicators such as "fear", "surprise", and "sadness", which indicates that "hospital" is normally related to the negative emotions of a character.  
The word ``however'' expresses a strong turning point, and it said that he failed the test.
The word "however" foretells a strong turning point and sad results that the character will fail the test in the hospital,
which is also consistent with the ``sadness'' and ``anger'' emotions that we input.

\subsection{Case Study}
\subsubsection{Comparison with Baseline Models}
Table \ref{examples1} shows the examples of stories generated by our model and the baseline models. Our model can generate a story that is coherent and consistent story with the psychological state we previously provided.

For the Seq2Seq model,  we see that it usually generates repeated sentences, and it does not consider all the characters. Example 1 in Table \ref{examples1}, generated by the baseline models describe only one character, ``Jane'', but, for the SoCP can generate information about the ``Friends''. In example 2, we see that the baseline models cannot vary the story with our given psychological state and even has wrong emotions, but our SoCP model can fit it properly. The GPT-2 model can generate reasonable sentences but has many repetitions.

Overall, our proposed model can generate stories by controlling characters' emotions. However, we see that not all stories are coherent; thus, the coherence is also a challenge confronting our model.

\subsubsection{Controllability Over Stories}
In Table \ref{examples2} and Table \ref{examples3},
we show examples of controllability generated stories under different conditions. 
The first example in Table \ref{examples2} compares the generated stories using an identical Plutchik indicator however under different scores.
In particular, in the first example, we assign the Plutchik indicator ``joy'' with different scores:
1, 0.5 and 0. When the score is equal to 1, it generates some obvious words, such as ``good'', ``happy'', and ``was proud of''. As the score of the `joy' is lower, the generated words are more and more negative. When the score is equal to 0, it generates some negative words, such as ``nervous'', ``tired'', and ``not happy''.
The second example shows generated stories with different Plutchik indicators. We assign ``surprise'', ``fear'', and ``anger'' as different Plutchik indicators. When Plutchik is ``surprise'', it generates some words, such as ``was surprised to'', and ``shocked''. When Plutchik indicates ``fear'', it generates some words, such as ``was afraid of'', and ``be scared to''. When Plutchik indicates ``anger'', it generates some words, such as ``angry''.
For the third example, we assign multiple Plutchik indicators of different scores. We can see the generated stories have multiple emotions.

In Table \ref{examples3}
, we show the different examples of controllable generated stories.
Similarly, they have an obvious effect.
According to the examples shown above, we see the controllability of our model.
There exist some stories that are not coherent in the above examples, which indicate our model pays more attention to the psychological states and overlooks the coherence of the story. This is where we need improvement.

\section{Conclusion}

Traditional story generation models show poor performance because they only take into account emotional changes in story development without considering the internal emotional and motivational changes of characters.
However, a story is usually written according to people's ideas and settings. Therefore, we have proposed a method that can specify the psychological changes in different characters in a story. In the application of the model, the psychological state can be assigned to the psychological robot so that it can generate helpful words for people with mental diseases. For future work, we will add an external knowledge graph to our model to better control word generation, and we will use an advanced transformer model to represent the text.


\end{document}